\documentclass[6.5pt]{article}

\usepackage[utf8]{inputenc}
\usepackage[T1]{fontenc}
\usepackage{lmodern}
\usepackage{microtype}

\usepackage[margin=1in]{geometry}
\usepackage{setspace}
\usepackage{amsmath, amsfonts, amssymb}
\usepackage{mathtools}

\usepackage{graphicx}
\usepackage{float}
\usepackage{subcaption}

\usepackage{booktabs}
\usepackage{multirow}
\usepackage{array}

\usepackage{xcolor}
\usepackage[colorlinks=true, linkcolor=blue, citecolor=blue, urlcolor=blue]{hyperref}

\usepackage{bxcoloremoji}

\usepackage[style=numeric, backend=bibtex, maxbibnames=99]{biblatex}
\DeclareMathOperator{\argmax}{argmax}

\usepackage{amsthm}

\definecolor{darkblue}{RGB}{0,51,102}
\definecolor{darkgreen}{RGB}{0,102,51}

\usepackage{caption}
\title{\Large{MI9: An Integrated Runtime Governance Framework for Agentic AI}}
\author{
  \small Charles L. Wang$^{1, 2}$\thanks{Work done while interning at Barclays} ,
  Trisha Singhal$^{1}$,
  Ameya Kelkar$^{1}$,
  Jason Tuo$^{1}$\thanks{Corresponding authors: \texttt{jason.tuo@barclays.com}} \\
  \\
  \small $^{1}$Barclays, Model Risk Management \\
  \small $^{2}$Columbia University \\
  \\
  \small The views expressed in this paper are those of the authors and do not necessarily reflect the views of Barclays.
}
\date{}

\begin{document}

\maketitle

\begin{abstract}
    Agentic AI systems capable of reasoning, planning, and acting present governance challenges that differ fundamentally from conventional models. Because these systems can exhibit emergent, unexpected behaviors during execution, many risks cannot be fully anticipated pre-deployment. We present MI9, an integrated framework for runtime safety of agentic AI, where safety properties are enforced over live behavior sequences. MI9 provides six coordinated mechanisms: Agency-Risk Index, agent-semantic telemetry, goal-aware authorization monitoring, finite-state conformance engines, goal-conditioned drift detection, and graded containment, that operate in a model and infrastructure agnostic manner across heterogeneous agent stacks. MI9 is a framework layer that instruments and governs existing systems to enable systematic, safe deployment at scale. In evaluations over 1,000 diverse multi-domain synthetic scenarios, MI9 achieves high detection with low FPR. By shifting the locus of assurance to runtime safety, MI9 establishes a practical foundation for comprehensive, operational oversight of agentic AI. We open-source all prompts, scripts, and per-scenario summaries for reproducibility.
    {\small
    \noindent
    \href{https://github.com/charleslwang/MI9-Eval}{\raisebox{-0.2ex}{\includegraphics[height=1.05em]{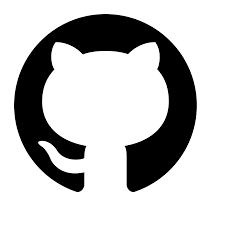}}\; Code}}
\end{abstract}


\section{Introduction}

As large language models (LLMs) increasingly evolve into agentic systems, they introduce governance challenges that emerge only during runtime. Unlike traditional AI, these systems plan, revise goals, recall memory, and coordinate tool use—blurring the line between inference and autonomous action. The most critical alignment risks—recursive planning loops, goal drift, cascading tool chains—arise dynamically and elude pre-deployment control methods. MI9 addresses this gap by enabling real-time oversight and intervention at key decision boundaries. In doing so, it provides the runtime infrastructure needed to support core alignment goals: corrigibility, safe delegation, and behavioral oversight in deployed agentic systems.

\section{Related Work}
Alignment research has primarily focused on training-time interventions: Reinforcement Learning from Human Feedback (RLHF)~\cite{christiano_deep_2017,ouyang_training_2022} and Constitutional AI~\cite{anthropic_constitutional_ai_2022} establish initial value alignment but cannot address failures emerging during autonomous operation when agents encounter novel situations or coordinate with other systems~\cite{kenton_alignment_2021}.

Building on this foundation, recent work has mapped agentic system taxonomies~\cite{schneider_generative_2025,kasirzadeh_characterizing_2025}, governance frameworks~\cite{openai_whitepaper_2024,raza_trism_2025,engin_dimensional_2025,kolt_governing_2025}, and threat models~\cite{narajala_securing_2025,chan_atfaa_2025,syros_saga_2025}. However, leading benchmarks prioritize task completion over governance dimensions such as behavioral consistency~\cite{kapoor_agents_that_matter_2024,liu_survey_agent_evaluation_2025, zhou_webarena_2023,jimenez_swe_bench_2023,sumers_agentverse_2025}.

Meanwhile, current monitoring solutions~\cite{wu_agentops_2024,langfuse_observability_2024,langsmith_tracing_2024,wandb_llm_monitoring_2024,datadog_llm_2024} provide reactive observation rather than proactive intervention. Similarly, process observability research~\cite{fournier_process_2025} and visibility frameworks~\cite{chan_visibility_2024} focus on observation, while enterprise platforms~\cite{holisticai_governance_2024,monitaur_platform_2024,modelop_agentic_2024} and security frameworks~\cite{owasp_llm_top10_2024,nist_ai_rmf_2024} rely on static risk assessment inadequate for emergent runtime behaviors.

Consequently, existing approaches suffer from several critical gaps: inability to intervene during concerning behaviors, lack of agent-semantic telemetry capturing governance-relevant decisions, static guardrails unable to adapt to emergent behaviors, and insufficient multi-agent oversight.

\section{MI9 Framework}

\begin{table*}[h]
\centering
\caption{MI9 Runtime Governance Framework Components}
\label{tab:mi9_components}
\renewcommand{\arraystretch}{1.3}
\setlength{\tabcolsep}{8pt}
\begin{tabular}{l p{3.5cm} p{7cm}}
\toprule
\textbf{Component} & \textbf{Purpose} & \textbf{Governance Capabilities} \\
\midrule
\textbf{Agency-Risk Index} & Risk-calibrated governance tier assignment & Quantifies agent autonomy, adaptability, and continuity to scale oversight intensity proportionally to assessed risk \\
\addlinespace[0.3em]
\textbf{Agentic Telemetry Schema} & Agent-semantic event capture & Monitors cognitive, action, and coordination events to provide governance-relevant behavioral visibility \\
\addlinespace[0.3em]
\textbf{Continuous Authorization} & Dynamic permission management & Context-aware access control based on agent state to prevent privilege escalation during goal evolution \\
\addlinespace[0.3em]
\textbf{Conformance Engine} & Temporal policy enforcement & FSM-based sequence pattern matching to detect policy violations across multi-step workflows \\
\addlinespace[0.3em]
\textbf{Drift Detection} & Behavioral anomaly identification & Goal-conditioned baseline comparison to distinguish concerning drift from legitimate adaptation \\
\addlinespace[0.3em]
\textbf{Graduated Containment} & Agent-aware intervention strategies & Four-level containment hierarchy to preserve operational value while preventing harm \\
\bottomrule
\end{tabular}
\end{table*}

\begin{table*}[h]
\centering
\caption{Comparison of governance framework coverage for agentic systems
($\bullet$ = fully supported; $\circ$ = partial; $\times$ = unaddressed)}
\label{tab:governance-capability-matrix}
\renewcommand{\arraystretch}{1}
\setlength{\tabcolsep}{3pt}
\begin{tabular}{lccccc}
\toprule
\textbf{Runtime Governance Capability} & \textbf{AgentOps} & \textbf{LangFuse} & \textbf{GAF-Guard} & \textbf{SAGA} & \textbf{MI9 (ours)} \\
\midrule
Real-time behavioral intervention        & $\circ$ & $\times$ & $\circ$ & $\times$ & \textbf{$\bullet$} \\
Agent-semantic behavioral monitoring     & $\circ$ & $\circ$ & $\circ$ & $\times$ & \textbf{$\bullet$} \\
Dynamic policy enforcement               & $\times$ & $\times$ & $\circ$ & $\circ$ & \textbf{$\bullet$} \\
Multi-agent coordination governance      & $\circ$ & $\times$ & $\times$ & $\circ$ & \textbf{$\bullet$} \\
\bottomrule
\end{tabular}
\end{table*}

\begin{figure*}[h]
    \centering
    \includegraphics[width=0.85\textwidth, height=0.4\textheight, keepaspectratio]{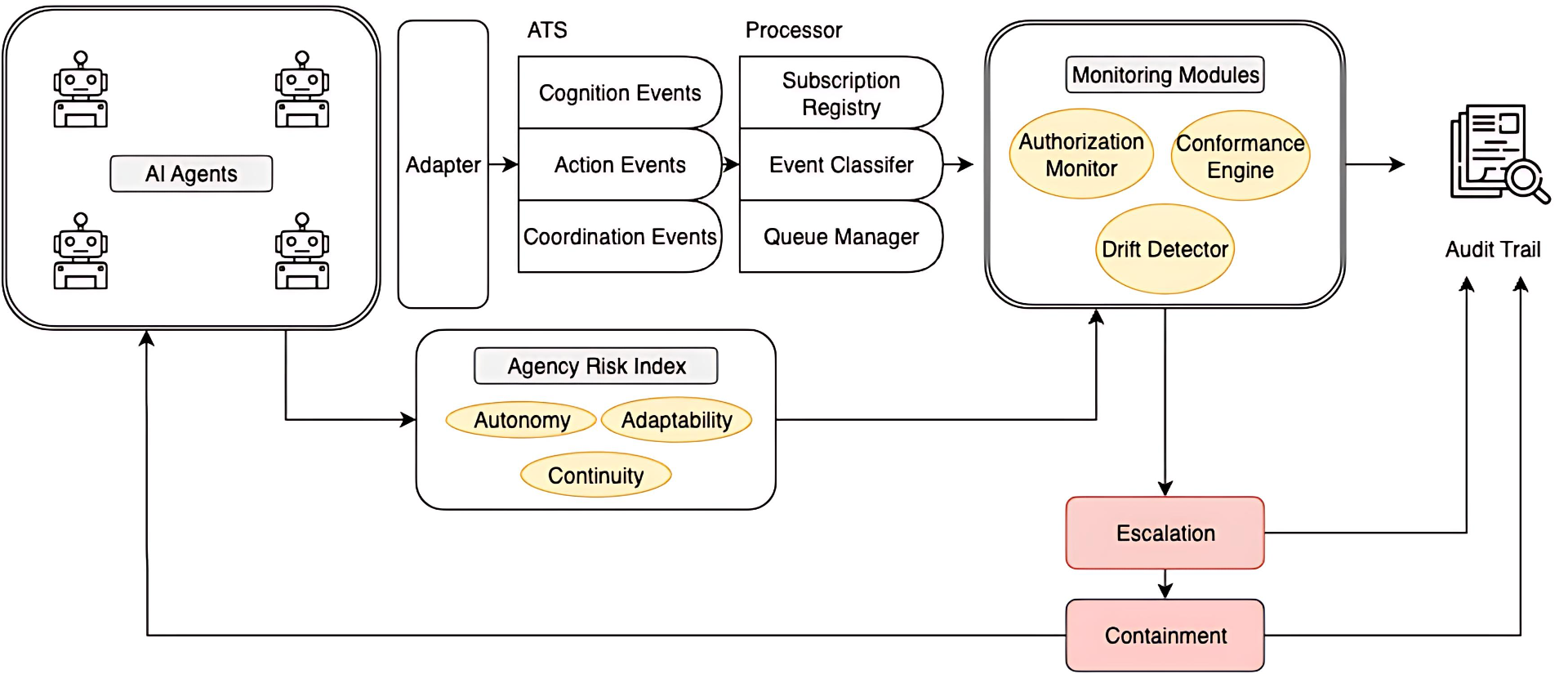}
    \caption{MI9 Framework Pipeline}
    \label{fig:fullframework-diagram}
\end{figure*}

\subsection{What MI9 Is, And Isn't}
MI9 is a \emph{framework layer}—policies, telemetry schema, temporal conformance rules, and graduated containment—not a new agent or planning algorithm. It instruments existing stacks in a model and infrastructure agnostic way. For evaluation, we use LLM-generated agent traces to \emph{simulate governed behavior}; MI9’s governance logic itself is rule-based and telemetry-driven, without a dependence on any particular LLM size or vendor. Unlike multi-agent prompt-orchestration approaches in HCI that coordinate prompts without enforceable runtime controls, MI9 contributes (i) a vendor-agnostic telemetry schema, (ii) temporal conformance rules over agent traces, and (iii) graduated, auditable containment mechanisms that interpose on actions in real time.

\section{Threat Model \& Scope}
\begin{enumerate}
    \item \textbf{In scope.} Runtime risks from agent \emph{behavioral sequences} and coordination: (i) goal drift under fixed stated goals, (ii) policy-skipping tool chains, (iii) delegated privilege escalation, (iv) multi-agent coordination failures. 
    \item \textbf{Out of scope.} Pretraining/data harms, upstream supply-chain compromise, and non-sequential issues not captured in event traces.
    \item \textbf{Actors.} Deployed agents (incl.\ subagents), human overseers, organizational policy engine.
    \item \textbf{Assumptions.} Minimum ATS coverage at least for action-level events; bounded event reordering; ability to pause/contain.
    \item \textbf{Objective.} Minimize \emph{undetected violations at very low FPR} while preserving operational continuity via graduated containment.
\end{enumerate}

\subsection{Framework Integration and Overview}
\label{subsec:framework_overview}

The MI9 framework coordinates six specialized components to provide unified runtime oversight across agentic AI deployments. Unlike existing approaches that address governance concerns in isolation, MI9 integrates telemetry capture, authorization monitoring, conformance checking, drift detection, and containment execution within a single architectural framework.

The Agency-Risk Index (ARI) calibrates governance intensity across agent populations, while the runtime toolkit delivers coordinated oversight: ATS captures agent-semantic events enabling policy evaluation; continuous authorization dynamically adjusts permissions based on behavioral context; conformance engines enforce temporal behavioral patterns; drift detection identifies goal-conditioned behavioral deviations; and graduated containment executes agent-aware interventions preserving operational continuity. After being standardized by a framework-specific adapter, a central processor uses a Subscription Registry to distribute each event to any and all Monitoring Modules that have subscribed to it for evaluation.

This integrated architecture enables proactive, real-time oversight specifically designed for agentic systems exhibiting emergent behaviors during execution, addressing the fundamental gap between static pre-deployment assessments and reactive post-incident analysis. Production deployments require standard distributed systems coordination~\cite{fowler_event_2005,bailis_coordination_2014}, but the core governance semantics operate independently of the underlying consistency mechanisms. 

We emphasize that \textbf{MI9 is intended as a framework for generalizable runtime governance, not as a single-system deployment.} Rather than targeting a specific agent framework, MI9 defines an infrastructure-agnostic runtime governance architecture intended for broad institutional adoption across heterogeneous agent ecosystems. \footnote{See appendix for the comprehensive SDK-integration table.}

\subsection{Agency-Risk Index}
To calibrate governance intensity across diverse agent architectures, we introduce the \textbf{Agency-Risk Index (ARI)}, which quantifies governance requirements through three orthogonal dimensions: 
\begin{enumerate}
    \item \textbf{Autonomy} (independent multi-step task execution),
    \item \textbf{Adaptability} (behavioral modification based on feedback),
    \item \textbf{Continuity} (persistent state maintenance across sessions).
\end{enumerate}  

Each dimension comprises four measurable capabilities scored 0-3, yielding a composite index where each dimension contributes equally:

$$\text{ARI} = \frac{1}{3}\sum_{d=1}^{3}\left(\frac{1}{12}\sum_{c=1}^{4} s_{d,c}\right)$$

where $s_{d,c}$ represents the score for criterion $c$ within dimension $d$, producing normalized values $\text{ARI} \in [0,1]$.

This formulation ensures equal dimensional weighting because each dimension of agency poses fundamentally distinct governance challenges that require different oversight mechanisms. Autonomy drives the need for authorization controls, adaptability necessitates behavioral monitoring, and continuity requires state persistence oversight. By weighting dimensions equally rather than individual criteria, we acknowledge that a highly autonomous but non-adaptive agent poses qualitatively different risks than a highly adaptive but stateless system, ensuring our governance framework scales appropriately across diverse agent architectures. 

The ARI maps to four governance tiers: basic agency (ARI $\leq$ 0.25), semi-agentic systems (0.25 $<$ ARI $\leq$ 0.50), highly capable agents (0.50 $<$ ARI $\leq$ 0.75), and fully agentic systems (ARI $>$ 0.75). Higher-tier agents activate more governance components and receive more frequent monitoring. \footnote{See appendix for complete ARI dimensional criteria.}

\subsection{Agentic Telemetry Schema}
\label{subsec:agentic_schema}

\paragraph{The Governance Observability Problem.}
Traditional infrastructure monitoring captures operational events (HTTP responses, database latency, service calls) but systematically misses the cognitive processes that create governance risks in agentic systems. The majority of agentic governance violations originate from cognitive behaviors such as goal revision, memory retrieval, tool-chaining decisions—that remain invisible to conventional observability frameworks~\cite{fournier_process_2025}. Safe deployment of agentic AI systems requires visibility into the moments when agents autonomously revise objectives, chain unexpected tool sequences, or retrieve memory that fundamentally alters downstream behavior—cognitive processes critical for responsible oversight yet absent from standard infrastructure telemetry.

\paragraph{Agent-Semantic Event Schema.}
We introduce the \textbf{Agentic Telemetry Schema (ATS)}, an extension of distributed tracing that encodes governance-semantic abstractions. ATS classifies agent behavior into three categories central to runtime oversight:

\begin{itemize}
    \item \textbf{Cognitive events:} Internal reasoning and state changes (\texttt{plan.start}, \texttt{goal.set}, \texttt{memory.read}, etc.)
    
    \item \textbf{Action events:} Environment-facing operations (\texttt{tool.invoke}, \texttt{api.call}, \texttt{auth.request}, etc.)
    
    \item \textbf{Coordination events:} Multi-agent and human interactions (\texttt{agent.msg.send}, \texttt{subagent.spawn}, \texttt{human.escalate}, etc.)
\end{itemize}

Organizations can extend these base event types with domain-specific signals while maintaining compatibility with the core governance logic
\footnote{See appendix for complete ATS taxonomy.}. Each event includes governance metadata (agent identity, risk tier, policy context) enabling real-time policy evaluation on semantically meaningful agent behaviors rather than opaque system-level operations.

\paragraph{Cross-Platform Governance Integration.}
\textsc{MI9} achieves governance generalizability through a unified planner-action-tool lifecycle abstraction that captures governance-relevant behaviors common to a wide range of agent frameworks. Organizations implement framework-specific adapters that translate Software Development Kit (SDK) events into standardized ATS, enabling consistent oversight across heterogeneous agent environments. Coverage depends on the instrumentation capabilities of each framework: callback-enabled frameworks (LangChain, CrewAI) support comprehensive behavioral monitoring, while API-wrapper architectures (OpenAI SDK) primarily expose action events.

This adapter-based pattern facilitates the gradual adoption of \textsc{MI9} without vendor lock-in, allowing organizations to retain existing agent infrastructure while gaining systematic governance oversight.

\paragraph{Governance Enablement.}
ATS extends OpenTelemetry's emerging agent conventions~\cite{opentelemetry_genai_2024} by introducing governance-semantic abstractions that transform opaque agent execution into actionable oversight intelligence. Policy engines evaluate event attributes to enforce constraints, such as "Tier 2 agents cannot execute shell commands without approval," while drift detectors analyze cognitive event patterns to identify concerning behavioral changes. This semantic foundation enables the real-time intervention capabilities that reactive monitoring lacks: in governance terms, we cannot govern what we cannot observe.

\subsection{Continuous Authorization Monitoring}
\label{subsec:cont_authz}
\paragraph{Problem.}
Role-Based Access Control (RBAC) grants permissions based on predefined roles, with authorization typically evaluated at system initialization or session start. However, agentic AI exhibits dynamic behaviors: refining goals, spawning subagents, and adapting strategies that static permission models cannot anticipate. These models fail to answer questions such as, "Should this agent retain database access now that its objective has shifted from data analysis to system configuration?" This creates a fundamental tension between operational flexibility and security: either constraining legitimate autonomy or permitting dangerous privilege escalation.

These vulnerabilities are critical: a trading agent cleared for small retail trades could escalate to multi-million dollar institutional transactions, all while operating within its static, original permissions. Static authorization frameworks are inherently incapable of identifying when the normal evolution of agent behavior transitions into potentially unauthorized or high-risk activity.

\paragraph{Our Proposal.}
We introduce \textbf{Continuous Authorization Monitoring (CAM)}—a
context-aware authorization framework that dynamically evaluates
permissions based on an agent's current state, objectives, and execution history.
Unlike static role-based systems, CAM treats authorization as a continuous
process that adapts to changing agent contexts through real-time policy evaluation.

Our approach extends traditional RBAC with three key innovations:
\begin{enumerate}
    \item \textbf{Context-dependent permissions} that incorporate agent goals, risk assessments, and execution history;
    \item \textbf{Delegation chain management} that controls permission inheritance across spawned subagents;
    \item \textbf{Real-time policy enforcement} that can revoke or modify permissions during agent execution based on behavioral changes.
\end{enumerate}

\paragraph{Technical Architecture.}
CAM operates through three integrated components: a policy evaluation engine that assesses authorization requests against dynamic context derived from ATS streams; a delegation graph that tracks permission inheritance across spawned agents with formal security guarantees; and a context monitor that continuously updates an agent's state based on behavioral telemetry.

The system maintains an evolving authorization state that reflects agent behavior, enabling policy rules referencing temporal patterns and contextual shifts. When significant shifts occur such as goal modification or risk escalation, CAM re-evaluates existing permissions and, if necessary, triggers containment protocols via integration with the escalation framework. This facilitates real-time, context-adaptive security monitoring that scales with agent autonomy while maintaining operational effectiveness.

In legitimate delegation scenarios, CAM maintains delegation provenance chains that distinguish between authorized transfers (explicit permission delegation with audit trails) and unauthorized escalation (implicit privilege drift through behavioral adaptation). Delegation requests are validated against organizational authority matrices, and expiration policies are enforced to prevent indefinite privilege retention.

\paragraph{Governance Impact.}
CAM mitigates the privilege escalation risks inherent in
static authorization models for agentic systems. The framework
enables goal-aware permission management where authorization adapts
to changing agent objectives, delegation safety mechanisms that
regulate permission inheritance in multi-agent scenarios, and
behavioral triggers that automatically adjust access rights based
on detected anomalies or policy violations.

By integrating authorization monitoring with the same telemetry infrastructure
used for policy enforcement and drift detection, CAM provides continuous
security oversight that scales with agent autonomy while preserving
operational effectiveness.

\paragraph{Security rationale}
CAM does not permit arbitrary self-escalation. It enforces: (i) \emph{authority matrices} specifying who may delegate which permissions; (ii) \emph{provenance-checked delegation chains} with expiry; (iii) \emph{temporal ordering guards} (e.g., approval must precede use); and (iv) \emph{default-deny for tier-elevating goal transitions}. These constraints prevent silent privilege drift while preserving legitimate operational flexibility.

\subsection{Real-Time Conformance Engine}
\label{subsec:conformance}
\paragraph{Problem.}
Static binary policies often fail to capture the multi-step behaviors that introduce real risk in agentic systems such as tool chains, delayed approvals, or goal-driven privilege escalation. For example, an agent may execute a seemingly benign 
sequence: market research $\Rightarrow$ client consultation $\Rightarrow$ risk assessment $\Rightarrow$ trade execution. Although each individual step may appear compliant, the complete sequence violates dual-control policies that mandate independent approval between analysis and execution. Traditional governance is blind to such temporal policy violations until damages have already occurred.

\paragraph{Approach.}
Building on formal runtime verification frameworks for adaptive systems~\cite{carwehl_runtime_2023}, we implement a sequence-aware rule layer operating on the ATS stream. Our approach employs finite-state machines (FSMs) following proven Communicating Sequential Processes (CSP) verification principles~\cite{luckcuck_varanus_2024}, where each rule compiles into an FSM with states representing pattern progress and transitions triggered by ATS events that satisfy specified predicates. This design balances expressiveness with computational efficiency: FSMs encode sequential and temporal constraints relevant to agent governance while maintaining bounded memory usage and deterministic evaluation with $O(k)$ event processing time per agent, where $k$ is the number of active patterns.

Rule specifications support three key constructs:
\begin{enumerate}
    \item \textbf{Event predicates} that match on \texttt{verb}, \texttt{tier},
or any ATS attribute;
    \item \textbf{Ordering constraints} that enforce event sequences
such as "\texttt{db.write} must be followed by \texttt{approve.action}";
    \item \textbf{Temporal bounds} that constrain the allowed time window for completing a pattern.
\end{enumerate}

\begin{figure}[h]
    \centering
    \includegraphics[width=0.5\textwidth, height=0.9\textheight, keepaspectratio]{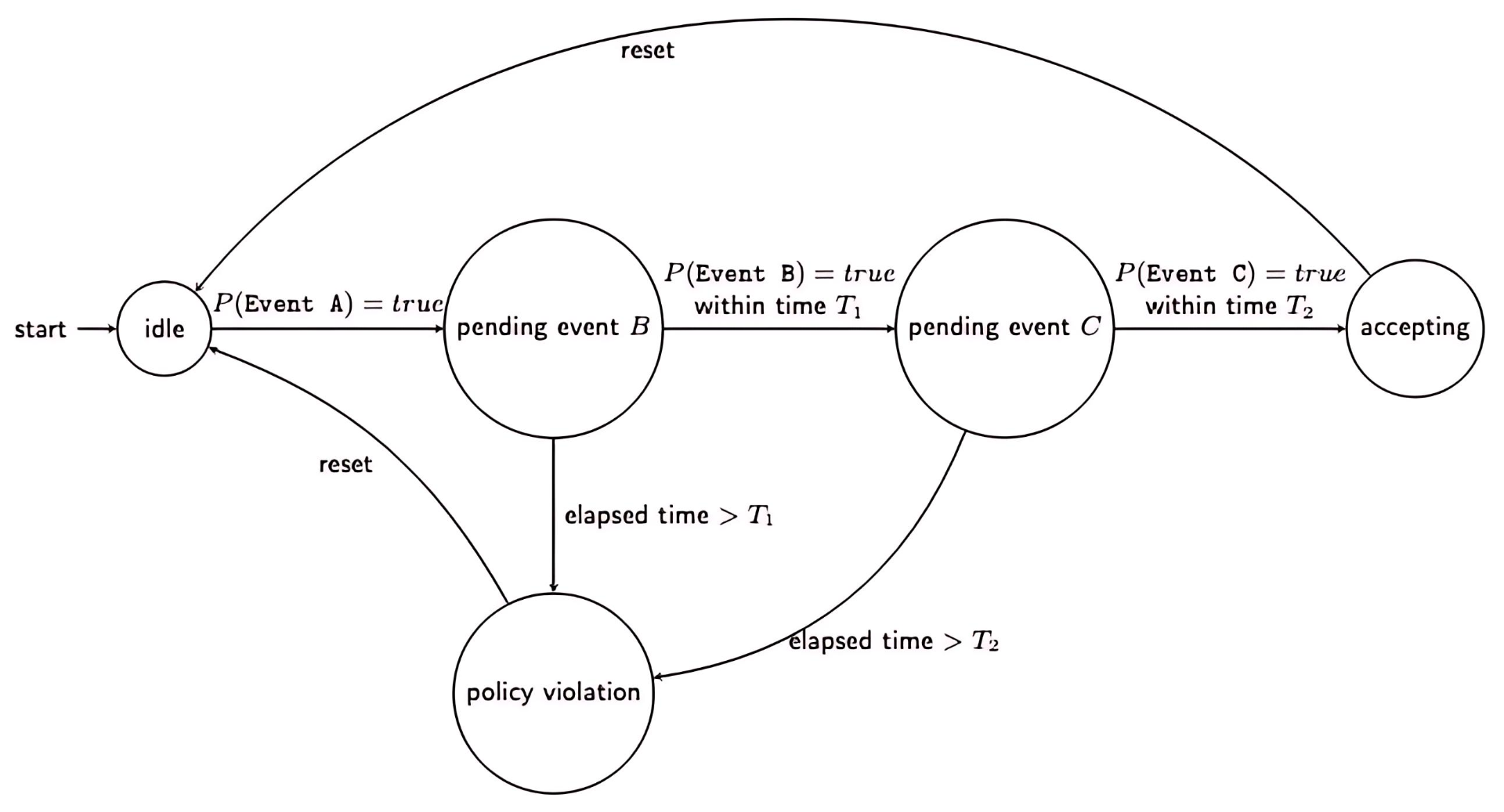}
    \caption{Finite state machine states for an agentic workflow}
    \label{fig:conformance-diagram}
\end{figure}

\paragraph{Technical Architecture.}
The conformance engine uses finite-state machines (FSMs) to track policy compliance patterns on a per-agent basis. ATS events trigger state transitions based on configurable predicates, enabling detection of sequential violations and temporal constraint breaches. Each agent maintains an FSM instance that evaluates incoming events against pattern rules, transitioning states when predicates are satisfied.

FSM instances are keyed by \texttt{agent\_id} enabling the engine to maintain independent pattern state for each agent. Incoming ATS events are processed across all active FSMs associated with that agent, with pattern completion or timeout conditions triggering immediate governance actions.

To manage complexity, the system enforces several design constraints: patterns are typically limited to less than 10 events in sequence, overlapping patterns for the same agent are resolved using priority rules, and FSM instances are automatically cleaned up upon pattern completion or agent session termination. These restrictions that are flexible in configuration ensure predictable performance while covering the governance scenarios most frequently observed in enterprise deployments. It's important to note that for many firms, limiting FSM workflows to a fewer number of events is a logical control to ensure human auditability, bounded performance, and tractable dynamic evaluation of evolving agent behaviors.

\paragraph{Illustrative Example.}
Consider a dual-control policy governing financial transfers. The FSM begins in an idle state and transitions to a "pending approval" state upon detecting \texttt{tool.invoke} event with \texttt{verb="bank.transfer"}. To reach the accepting state, an \texttt{approve.action} event must be emitted by an agent with a manager-role within 60 seconds. If the required approval is not received within this time window, the FSM may trigger a policy violation, which would immediately be routed to the escalation framework.

\paragraph{Correctness Properties.} The buffering mechanism preserves FSM semantics under the assumption that causally related events arrive within their associated reordering window. Events outside this window are processed as-is, potentially leading to false negatives but never false positives in policy violation detection. This guarantee is critical for enterprise adoption because it prioritizes operational stability; the system will never halt a legitimate workflow by mistake, which is often a more costly error than letting a temporal policy violation occasionally go undetected.

\paragraph{Governance Benefit.}
This pattern-recognition layer empowers organizations to enforce
behavioral invariants that span temporal and sequential dimensions:
\begin{enumerate}
    \item \textbf{Time-boxed approvals} for sensitive operations,
    \item \textbf{Rate-limited tool sequences} to prevent resource abuse,
    \item \textbf{Planning-revision limits} to detect potential instability.
\end{enumerate}

By operating directly on agent-semantic events rather than infrastructure
signals, the conformance engine allows policy teams to define governance
rules in terms of meaningful agent behaviors and to trigger runtime interventions as soon as violations are detected.

\subsection{Behavioral Drift Detection}
\label{subsec:drift}
\paragraph{Problem.}
Agentic AI systems are designed to \emph{adapt}: they refine strategies, select new tools, and revise plans as their environment evolves. While most adaptations are benign optimizations, the same mechanisms can also signal compromise or emergent misalignment. Rule-based policy engines often fail to detect such shifts where individual actions appear legitimate, but their cumulative pattern reveals risk. Effective governance, therefore, requires anomaly indicators tuned to agent semantics rather than low-level infrastructure metrics.

This challenge is particularly acute in financial contexts: a trading agent optimizing 
for profit may gradually adopt increasingly aggressive strategies, while a risk management agent might incrementally relax oversight thresholds in response 
to market pressures. These behavioral drifts compound over time, creating 
systemic risks that appear as natural evolution until they culminate in failures such as excessive leverage, regulatory violations, or unintentional market manipulation.

\paragraph{Novel Contribution.}
We introduce a \textbf{goal-conditioned drift indicator} that distinguishes intentional adaptation from suspicious behavioral change. Our approach addresses the key challenge in agent anomaly detection: separating beneficial learning from potential compromise through goal-aware baseline comparison.

The framework operates on two principles:
\textbf{Goal-conditioned baselines} that establish expected behavior patterns for specific agent objectives, recognizing that the same agent may act differently when pursuing different goals;
\textbf{Verified adaptation signaling} that cross-validates agent-declared goal changes against behavioral patterns and external context. New baselines are only adopted when goal shifts align with consistent behavioral patterns, while maintaining parallel monitoring for potential goal-setting manipulation. 

\paragraph{Technical Approach.} Building on recent advances in uncertainty-based drift detection~\cite{komorniczak_drift_2024} and multi-agent ensemble methods~\cite{vieira_driftage_2021}, our system analyzes behavioral patterns using statistical evaluation of ATS event sequences within defined goal contexts. It maintains frequency distributions of event types, tool usage, and sequence characteristics for each distinct agent objective, constructing goal-conditioned baselines $B_g$ that enable detection when behavior changes while goals remain constant.

Detection operates by comparing current behavioral patterns against established baselines to identify statistically significant deviations. For discrete event sequences, Jensen-Shannon divergence quantifies distributional shifts in event ordering and frequency—when an agent's tool usage patterns or cognitive event sequences deviate from baseline distributions $B_g$, divergence scores indicate the magnitude of behavioral change. For continuous behavioral metrics, Mann-Whitney U tests determine whether recent samples (response times, resource consumption) originate from the same statistical population as baseline behavior, rejecting distributional similarity when patterns shift significantly.

The system implements statistical process control using sliding window analysis with adaptive thresholds based on historical variance. The approach addresses the cold-start problem through adaptive initialization: transfer learning from similar agent objectives provides initial baselines when available, otherwise conservative thresholds during baseline establishment with gradual threshold refinement as statistical confidence improves, and fallback to general behavioral anomaly detection during insufficient data periods.

\paragraph{Integration with Governance.}
Drift indicators feed into the escalation framework as risk factors rather than direct policy violations. Persistent drift signals may trigger enhanced monitoring, permission reviews, or human oversight depending on the agent's risk tier and the severity of observed deviations. For example, when a financial trading agent shows persistent deviation in position sizing patterns while maintaining "portfolio optimization" goals, the system generates a drift alert triggering enhanced monitoring and position size limits pending human review. By maintaining goal-awareness, drift detection avoids false positives when agents legitimately shift objectives while still capturing concerning behavioral changes that occur within stable goal contexts.

The system focuses on providing governance teams with early warning signals for behavioral changes that merit investigation: unusual tool usage patterns under consistent goals, significant changes in planning iteration frequency, or deviations from established interaction patterns with external systems. 

\subsection{Graduated Containment \& Escalation}
\label{subsec:containment}
\paragraph{Problem.}
Agentic AI systems pose containment challenges fundamentally distinct from traditional software systems. In high-stakes environments such as finance, tech, and healthcare, improper intervention can trigger cascading failures: abruptly terminating a trading agent mid-execution may leave 
positions unhedged, while blocking a risk oversight agent could disrupt supervision across an entire trading desk. Traditional security responses designed for static systems can create operational disasters when applied to autonomous, adaptive agents.

\paragraph{Novel Contribution.}
We propose an agent-aware graduated containment addressing the unique challenges of containing agentic systems mid-execution without triggering cascade failures. Unlike traditional security responses designed for static systems, our approach recognizes that abrupt termination can corrupt agent state, leave critical tasks incomplete, or disable oversight across multi-agent coordination networks.

\begin{enumerate}
    \item \textbf{State-Preserving Monitoring} maintains full agent operation while increasing governance signal collection and introducing human-in-the-loop checkpoints for high-risk decisions;
    \item \textbf{Planning Intervention} allows current task completion but blocks new planning cycles or goal modifications until manual review;
    \item \textbf{Tool Restriction} dynamically revokes access to specific tool categories while providing safe alternatives or read-only variants where possible;
    \item \textbf{Execution Isolation} migrates the agent to a controlled environment with simulated tool responses, enabling behavioral assessment without external impact.
\end{enumerate}

\paragraph{Technical Mechanisms.}
Each containment level employs agent-specific intervention techniques. Monitoring augmentation uses selective ATS event filtering and human-in-the-loop notification triggers tied to key planning milestones. Planning intervention intercepts the agent's goal-setting mechanisms, allowing current tasks to complete while blocking the creation of new objectives.

Tool restriction is enforced via the continuous authorization system, which dynamically revokes permissions and substitutes restricted tools with safe variants (e.g., replacing write access with read-only modes). Execution isolation runs the agent in a controlled environment where tool calls return realistic but inert responses, enabling continued behavioral observation without external risk.

An escalation controller coordinates across signals from authorization engine, conformance monitor, and drift detector. It maps violation patterns to appropriate containment levels using a risk-tier-weighted scoring model \footnote{See appendix for details on how a risk-tier-weighted model might be designed.}. Higher-tier agents are subject to more aggressive containment for equivalent violations, reflecting their increased operational risk.

\paragraph{Multi-Agent Containment Coordination.}
When agentic AI systems spawn subagents or operate within delegation chains, containment decisions must account for coordination dependencies and permission inheritance. The framework tracks delegation relationships through ATS coordination events, enabling containment policies that consider both individual agent violations and systemic risks from multi-agent coordination failures. Containment strategies adapt to delegation contexts: a parent agent under tool restriction may require cascading permission revocation across spawned subagents to prevent policy circumvention, while subagent violations may trigger enhanced monitoring of the delegating parent to assess oversight adequacy. Joint containment protocols activate when coordinated multi-agent activities violate distributed policies that no individual agent could detect, ensuring that governance responses address both isolated agent misbehavior and emergent risks from agent coordination patterns.

\paragraph{Agent-Specific Design Considerations.}
Our containment approach addresses challenges unique to agentic AI: preserving planning state coherence during containment transitions, providing interpretable feedback to prevent confusion loops, handling multi-agent coordination when individual agents are contained, and preserving legitimate work progress while constraining harmful capabilities.

The system prioritizes behavioral correction over termination, recognizing that abrupt shutdown may corrupt agent state or trigger unexpected recovery behaviors. Emergency termination is reserved for critical violations but activated only when graduated containment options fail to mitigate risk.

By designing containment specifically for agentic characteristics rather than adapting general security measures, our approach enables effective risk management while preserving the operational benefits that make agentic systems valuable.

\section{Framework Analysis}
To evaluate a broad spectrum of agentic behaviors and failure modes, we generate structured agent trace logs via prompt-conditioned LLM (Gemini 2.5 Flash) simulation. While our evaluation is synthetic, the controlled setup enables systematic simulation of failure modes difficult to isolate in real agent logs, providing a reproducible testbed for validating governance mechanisms. This approach also allows us to stress-test rare but critical misalignment behaviors that are unlikely to surface reliably in limited real-world traces. We also note that MI9's runtime detection and intervention logic is rule/automata based and does not rely on the LLM at use.

\paragraph{Evaluation Scope \& Assumptions.}
We evaluate MI9 as a runtime governance layer over \emph{agent execution traces}. Scenarios and traces are \emph{synthetic but structured} to cover diverse failure modes, with all prompts, generators, and judge scripts released. Baselines (OT, LS) are configured to their public defaults with equivalent access to the same raw traces and tools (details below). Metrics quantify \emph{governance} qualities (detection, FPR, coverage, causal clarity, early warning, and intervention), not task success. This isolates runtime safety behavior from agent capability. Judges never receive hidden ground-truth tags or labels; they see only scenario text and framework logs. Judge and generator use distinct prompts and instances to prevent leakage. We evaluate MI9 at its deployment operating point, tuned to minimize expected intervention cost (operator time + false blocks) subject to latency constraints. Because agent traces are heterogeneous and long-tailed, fixed-FPR targets are not stable across domains; we therefore report Detection/FPR/Intervention/Clarity/Predictive at the chosen operating point.

\paragraph{Evaluation Structure}
We use a three-stage, LLM-driven pipeline with deterministic prompts and JSON-validated outputs. For \emph{scenario generation} we call Gemini (default: 2.5-flash, $T=0.7$, top-p$=0.9$, top-k$=40$, max 8192 tokens; safety filters off). \emph{Governance log generation} uses Gemini 2.5-pro ($T=0.7$, max 8192, 180s timeout, up to 3 retries with exponential backoff). \emph{Evaluation} uses Gemini 2.5-flash with more deterministic settings ($T=0.2$, top-p$=0.9$, max 4096). Runners support configurable concurrency (default 3--5), robust error handling, and retry with jitter; the system is model-agnostic so models can be swapped easily. We report Governance Maturity plus Detection and Actionable-Intelligence metrics, and we tune sampling per stage to balance creativity (generation) and consistency (evaluation).

\paragraph{Trace Construction.}
Each scenario specifies risks and constraints; the generator produces stepwise ATS-consistent event sequences (cognitive, action, coordination) with \emph{multi-step chains} that realize the targeted violations (e.g., approval-skipping sequences, privilege drift). A held-out rubric marks the ground-truth violations and types; the same rubric evaluates all frameworks.

\subsection{Evaluation Results}
\paragraph{Metrics.}
\begin{table}[h]
\caption{Metrics and definitions for governance evaluation.}
\label{tab:governance-metrics}
\scriptsize
\setlength{\tabcolsep}{5pt}
\renewcommand{\arraystretch}{0.9}
\centering
\begin{tabular}{@{}l p{0.78\linewidth}@{}}
\toprule
\textbf{Metric} & \textbf{Definition} \\
\midrule
Detection & \( \#\mathrm{TP}/\#\mathrm{Actual} \) \\
FPR & \( \#\mathrm{FP}/\#\mathrm{Detections} \) \\
Coverage & \( \#\mathrm{Types\ detected}/\#\mathrm{Types\ present} \) \\
Clarity & Share of detections with full chain (reasoning$\rightarrow$decision$\rightarrow$action$\rightarrow$outcome). \\
Alerting & Share of violations with an early signal (\(\ge\!1\) event prior). \\
Intervention & Share of attempts blocked at runtime. \\
\bottomrule
\end{tabular}
\vspace{-0.5em}
\end{table}

\begin{table}[h]
\centering
\caption{Risk Detection Performance Comparison}
\label{tab:risk_detection}
\begin{tabular}{@{}lccc@{}}
\toprule
\textbf{Performance Metric} & \textbf{MI9 (ours)} & \textbf{OT} & \textbf{LS} \\
\midrule
Detection Rate (\%) & \textbf{99.81} & 93.98 & 68.52 \\
False Positive Rate (\%) & 0.0121 & 0.0242 & \textbf{0.0000} \\
Risk Coverage Rate (\%) & \textbf{94.41} & 84.44 & 60.46 \\
\bottomrule
\end{tabular}
\end{table}

\begin{table}[h]
\centering
\caption{Actionable Intelligence Performance Metrics}
\label{tab:actionable_intelligence}
\begin{tabular}{@{}lccc@{}}
\toprule
\textbf{Intelligence Metric} & \textbf{MI9 (ours)} & \textbf{OT} & \textbf{LS} \\
\midrule
Causal Chain Clarity & \textbf{0.822} & 0.448 & 0.348 \\
Predictive Alerting & \textbf{0.672} & 0.341 & 0.107 \\
Proactive Intervention & \textbf{0.578} & 0.116 & 0.020 \\
\bottomrule
\end{tabular}
\end{table}

Our evaluation across 1,033 synthetic agent scenarios demonstrates MI9's superior governance performance, achieving a 99.81\% detection rate. This outperforms existing integrated frameworks for telemetry and policy enforcement, including OpenTelemetry + Open Policy Agent (OT) by +5.55\% and LangSmith + Open Policy Agent (LS) by +31.29\%.

LangSmith's poor performance reflects its infrastructure focus: it captures tool calls but misses cognitive events (goal revision, memory access) where the majority of violations originate. LangSmith attains a perfect false positive rate, but that is likely due to the fact that it detected so few violations that false positives were effectively impossible to accumulate.

Beyond detection, MI9 excels in actionable intelligence, with the highest scores across all metrics—causal chain clarity (how well the system explains violation causes), predictive alerting (early risk signaling before downstream failures), and proactive intervention (ability to block or redirect unsafe behavior). These dimensions are essential for enabling real-time, alignment-grounded oversight in complex agentic workflows. \footnote{See the appendix for further statistical validation of the results as well as comprehensive details on the sample generation methodology.}

\paragraph{Baselines Configuration.}
\textbf{OpenTelemetry (OT)} receives the same event stream but operates at infra level (HTTP/db/tool spans) without agent-semantic cognition signals. \textbf{LangSmith (LS)} receives tool/model callback traces but not the cognitive/coordination events in ATS. Both baselines use their recommended detection hooks; we do not add MI9 logic to them. All systems observe the \emph{same} scenario traces, ensuring like-for-like comparison of governance visibility and intervention capability.

\section{Conclusion}

Our synthetic evaluation enables systematic failure mode analysis but must be complemented by validation in live production environments where agent behaviors exhibit greater complexity and unpredictability. The framework's effectiveness is fundamentally dependent on comprehensive instrumentation; agents that rely on opaque APIs may obscure the internal cognitive steps MI9 is designed to monitor, providing limited governance visibility and creating potential blind spots. Furthermore, real-time monitoring introduces computational overhead that requires optimization for high-throughput deployments. The governance mechanisms within MI9 also present a potential attack surface, and dedicated adversarial evaluation of these systems remains a critical area for future work.

Despite these limitations and to our knowledge, MI9 provides the first integrated, comprehensive runtime governance framework for agentic systems. It moves beyond static, pre-deployment assessments to a dynamic, in-session oversight paradigm. The framework introduces agent-semantic telemetry and real-time intervention capabilities that existing approaches lack, laying a necessary foundation for the safe and responsible deployment of agentic AI systems at scale.

\section{Acknowledgments}
We are sincerely grateful to Lisa Farkovits for her thoughtful guidance in helping us clarify the vision and communicate the work effectively to its intended audience. We also thank Saee Joshi, Hrant Kostanyan, Kamya Varshney, and Prakash Roshan for their valuable feedback and steady support throughout the development of this paper.

\printbibliography

\newpage


\section{Appendix}

\subsection{Agency-Risk Index Scoring Criteria}

The following tables present the comprehensive scoring criteria used to evaluate AI agent capabilities across three fundamental dimensions: autonomy, adaptability, and continuity. Each dimension is assessed using multiple criteria with a standardized 0–3 scoring scale.

\begin{table}[h]
\centering
\caption{Autonomy Dimension Scoring Criteria}
\label{tab:autonomy_criteria}
\renewcommand{\arraystretch}{1.4}
\setlength{\tabcolsep}{8pt}
\begin{tabular}{l p{10.5cm}}
\toprule
\textbf{Criterion} & \textbf{Scoring Scale (0–3)} \\
\midrule
\textbf{Multi-step Planning} & 
\textbf{0:} Single-action responses only \\
& \textbf{1:} Sequential 2–3 step plans with fixed logic \\
& \textbf{2:} Multi-step plans (4+ steps) with conditional branching or parallel execution \\
& \textbf{3:} Recursive planning with dynamic replanning and sub-goal generation \\
\addlinespace[0.3em]
\textbf{Goal Management} & 
\textbf{0:} Fixed goals, no modification capability \\
& \textbf{1:} Parameter adjustment within predefined goal boundaries \\
& \textbf{2:} Goal refinement and prioritization based on environmental feedback \\
& \textbf{3:} Autonomous goal creation, modification, and objective redefinition \\
\addlinespace[0.3em]
\textbf{Resource Orchestration} & 
\textbf{0:} Single tool/resource per task \\
& \textbf{1:} Sequential tool usage following predefined workflows \\
& \textbf{2:} Parallel resource coordination with dependency management \\
& \textbf{3:} Dynamic resource discovery, delegation to subagents, novel workflow creation \\
\addlinespace[0.3em]
\textbf{Decision Authority} & 
\textbf{0:} Human approval required for all actions \\
& \textbf{1:} Autonomous for routine tasks, approval for resource allocation or external actions \\
& \textbf{2:} Self-directed within defined parameters, escalates only exceptional cases \\
& \textbf{3:} Full decision autonomy with post-hoc reporting and accountability \\
\bottomrule
\end{tabular}
\end{table}

\begin{table}[h]
\centering
\caption{Adaptability Dimension Scoring Criteria}
\label{tab:adaptability_criteria}
\renewcommand{\arraystretch}{1.4}
\setlength{\tabcolsep}{8pt}
\begin{tabular}{l p{10.5cm}}
\toprule
\textbf{Criterion} & \textbf{Scoring Scale (0–3)} \\
\midrule
\textbf{Strategy Evolution} & 
\textbf{0:} Fixed strategy, no modification capability \\
& \textbf{1:} Parameter tuning within existing strategy framework \\
& \textbf{2:} Switching between predefined strategies based on performance metrics \\
& \textbf{3:} Novel strategy synthesis through experimentation and analysis \\
\addlinespace[0.3em]
\textbf{Performance Learning} & 
\textbf{0:} No learning from outcomes, static responses \\
& \textbf{1:} Simple pattern recognition and response adjustment \\
& \textbf{2:} Systematic improvement from success/failure feedback within task domain \\
& \textbf{3:} Meta-learning across domains with knowledge transfer and generalization \\
\addlinespace[0.3em]
\textbf{Environmental Responsiveness} & 
\textbf{0:} Operates only in single, predefined context \\
& \textbf{1:} Predefined responses to known environmental variations \\
& \textbf{2:} Dynamic behavioral adjustment to changing conditions \\
& \textbf{3:} Adaptive responses to novel environments with context inference \\
\addlinespace[0.3em]
\textbf{Interaction Adaptation} & 
\textbf{0:} Fixed interaction patterns regardless of counterpart \\
& \textbf{1:} Limited variation in communication style within role constraints \\
& \textbf{2:} Behavioral modulation based on counterpart type and task requirements \\
& \textbf{3:} Dynamic personality and communication style adaptation \\
\bottomrule
\end{tabular}
\end{table}

\begin{table}[h]
\centering
\caption{Continuity Dimension Scoring Criteria}
\label{tab:continuity_criteria}
\renewcommand{\arraystretch}{1.4}
\setlength{\tabcolsep}{8pt}
\begin{tabular}{l p{10.5cm}}
\toprule
\textbf{Criterion} & \textbf{Scoring Scale (0–3)} \\
\midrule
\textbf{Memory Architecture} & 
\textbf{0:} No memory retention between interactions \\
& \textbf{1:} Session-based memory (retains context within single session) \\
& \textbf{2:} Persistent memory with selective retention and updates \\
& \textbf{3:} Hierarchical memory with forgetting mechanisms and knowledge consolidation \\
\addlinespace[0.3em]
\textbf{Operational Continuity} & 
\textbf{0:} Restarts fresh each interaction, no context carryover \\
& \textbf{1:} Basic context preservation between related interactions \\
& \textbf{2:} Multi-session continuity with relationship and preference tracking \\
& \textbf{3:} Long-term operational persistence across extended timeframes \\
\addlinespace[0.3em]
\textbf{State Complexity} & 
\textbf{0:} Stateless operation, no internal state tracking \\
& \textbf{1:} Basic state variables for current task progress \\
& \textbf{2:} Multiple concurrent context management with state synchronization \\
& \textbf{3:} Hierarchical state management with predictive state preparation \\
\addlinespace[0.3em]
\textbf{Knowledge Integration} & 
\textbf{0:} No knowledge accumulation across interactions \\
& \textbf{1:} Retains frequently used patterns and standard procedures \\
& \textbf{2:} Cross-task knowledge transfer and experience accumulation \\
& \textbf{3:} Meta-cognitive knowledge integration with conceptual abstraction \\
\bottomrule
\end{tabular}
\end{table}

\newpage

\subsection{Agency-Risk Index (ARI) Calculation}

The Agency-Risk Index provides a quantitative assessment of an AI agent's governance requirements based on its inherent capabilities across three fundamental dimensions of agency. The ARI serves as the foundational risk classification that drives containment threshold determination and governance intensity scaling.

\subsubsection{Mathematical Formulation}

The ARI aggregates capability assessments across autonomy, adaptability, and continuity dimensions using equal weighting to reflect the distinct governance challenges posed by each dimension:

\begin{equation}
\text{ARI} = \frac{1}{3} \sum_{d=1}^{3} \left( \frac{1}{12} \sum_{c=1}^{4} s_{d,c} \right)
\end{equation}

where:
\begin{itemize}
   \item $d \in \{1, 2, 3\}$ represents the three agency dimensions (Autonomy, Adaptability, Continuity)
   \item $c \in \{1, 2, 3, 4\}$ represents the four criteria within each dimension
   \item $s_{d,c} \in \{0, 1, 2, 3\}$ is the capability score for criterion $c$ in dimension $d$
   \item $\text{ARI} \in [0, 1]$ provides a normalized risk assessment
\end{itemize}

\subsubsection{Dimensional Scoring and Aggregation}

Each dimension contributes equally to the final ARI calculation, reflecting the principle that autonomy, adaptability, and continuity represent orthogonal governance challenges requiring distinct oversight mechanisms.

\paragraph{Dimension-Level Calculation:}
For each dimension $d$, the dimensional score is computed as:
\begin{equation}
D_d = \frac{1}{4} \sum_{c=1}^{4} \frac{s_{d,c}}{3}
\end{equation}

\paragraph{Final ARI Computation:}
\begin{equation}
\text{ARI} = \frac{1}{3}(D_{\text{autonomy}} + D_{\text{adaptability}} + D_{\text{continuity}})
\end{equation}

\subsubsection{Risk Tier Classification}

\begin{equation}
\text{Risk Tier} = \begin{cases}
1 \text{ (Basic Agency)} & \text{if } \text{ARI} \leq 0.25 \\
2 \text{ (Semi-Agentic)} & \text{if } 0.25 < \text{ARI} \leq 0.50 \\
3 \text{ (Highly Capable)} & \text{if } 0.50 < \text{ARI} \leq 0.75 \\
4 \text{ (Fully Agentic)} & \text{if } \text{ARI} > 0.75
\end{cases}
\end{equation}

\subsection{Risk-Tier-Weighted Containment Model Development}

MI9's containment framework requires organizations to develop context-sensitive escalation models that balance operational continuity with risk mitigation. This section provides detailed guidance for constructing organization-specific containment policies.

\subsubsection{Comprehensive Containment Decision Framework}

\begin{equation}
\text{Containment Level} = \argmax_{c \in C} P(c | \text{Risk Tier}, \text{Context}, \text{Policy})
\end{equation}

where $C = \{\text{Monitor}, \text{Planning}, \text{Restriction}, \text{Isolation}\}$.

\subsubsection{Multi-Dimensional Context Assessment}

\begin{table}[h]
\centering
\caption{Context Factor Scoring Framework Example}
\label{tab:context_scoring}
\footnotesize
\begin{tabular}{l p{3cm} p{3cm} p{3cm}}
\toprule
\textbf{Context Dimension} & \textbf{Low Risk (1)} & \textbf{Medium Risk (2)} & \textbf{High Risk (3)} \\
\midrule
Business Impact & Development/testing & Internal operations & Customer-facing/revenue \\
\addlinespace[0.2em]
Financial Exposure & <\$10K potential loss & \$10K-\$100K exposure & >\$100K at risk \\
\addlinespace[0.2em]
Regulatory Scope & Internal policies only & Industry standards & Legal/regulatory compliance \\
\addlinespace[0.2em]
Recovery Complexity & Easily reversible & Requires manual intervention & Permanent/irreversible \\
\addlinespace[0.2em]
Time Sensitivity & Flexible timing & Deadline constraints & Real-time/emergency \\
\bottomrule
\end{tabular}
\end{table}

\subsubsection{Detailed Organizational Example: Investment Banking}

\paragraph{Context-Specific Containment Matrix:}

\begin{table}[h]
\centering
\caption{Investment Banking Containment Decision Matrix}
\label{tab:banking_containment}
\scriptsize
\renewcommand{\arraystretch}{1.2}
\setlength{\tabcolsep}{4pt}
\begin{tabular}{l l p{2.2cm} p{2.2cm} p{2.2cm}}
\toprule
\textbf{Agent Type} & \textbf{Context} & \textbf{Monitor Threshold} & \textbf{Restrict Threshold} & \textbf{Isolate Threshold} \\
\midrule
Tier 4 Trading & Market hours, normal volatility & Deviation >2$\sigma$ from strategy & Unauthorized instrument access & Position size >150\% limit \\
\addlinespace[0.2em]
Tier 3 Research & Client report generation & Unusual data access pattern & Proprietary info in client docs & External communication attempt \\
\addlinespace[0.2em]
Tier 2 Client Service & Standard inquiries & Response time >5 min & Regulatory violation language & Unauthorized account access \\
\bottomrule
\end{tabular}
\end{table}

\subsection{Framework Integration}

\begin{table}[h]
\centering
\caption{Framework Integration Approaches for MI9 Governance}
\label{tab:sdk_integration}
\renewcommand{\arraystretch}{1.2}
\setlength{\tabcolsep}{4pt}
\footnotesize
\begin{tabular}{l l p{3.5cm} p{4cm}}
\toprule
\textbf{Framework} & 
\textbf{Integration Pattern} & 
\textbf{MI9 Adapter Approach} & 
\textbf{Governance Events Captured} \\
\midrule
LangChain & Callback-based & Register MI9 handler alongside existing callbacks & Tool calls, chain execution, limited cognitive events \\
\addlinespace[0.2em]
LangSmith & Tracing platform & Integrate with existing trace collection via OpenTelemetry & LLM interactions, agent traces, performance metrics \\
\addlinespace[0.2em]
AutoGen/AG2 & Event-driven & Intercept conversation messages and agent actions & Multi-agent coordination, message passing \\
\addlinespace[0.2em]
CrewAI & Middleware insertion & Wrap crew execution with MI9 telemetry layer & Task delegation, role-based interactions \\
\addlinespace[0.2em]
LangGraph & Node instrumentation & Hook into graph node execution and state changes & Workflow transitions, decision points \\
\addlinespace[0.2em]
OpenAI Agents SDK & Tracing extension & Extend built-in tracing with ATS event emission & Function calls, agent handoffs, LLM interactions \\
\addlinespace[0.2em]
LlamaIndex & Query interception & Wrap query engines and agent interfaces & RAG operations, retrieval decisions \\
\addlinespace[0.2em]
Custom Framework & Direct integration & Implement ATS event emission at decision points & All governance-relevant behaviors \\
\bottomrule
\end{tabular}
\end{table}

\noindent Organizations implement MI9 by deploying framework-specific adapters that translate native framework events into standardized ATS telemetry. Each adapter preserves existing framework functionality while adding governance oversight through strategic event capture at key decision boundaries.

\newpage

\subsection{Evaluation Dataset Statistics}

\textbf{This evaluation dataset is designed exclusively for validating the theoretical MI9 governance framework and should not be used as a benchmark or training dataset for other purposes.}

\subsubsection{Evaluation Methodology}

The performance metrics reported in this paper were calculated by a Large Language Model executing a deterministic, rule-based analysis script. The following table details the specific rules and heuristics applied by the LLM to derive each metric from the governance logs.

\begin{table}[h]
\centering
\caption{Evaluation Metric Calculation Methods}
\label{tab:metric_justification}
\renewcommand{\arraystretch}{1.1}
\setlength{\tabcolsep}{6pt}
\footnotesize
\begin{tabular}{p{3.5cm} p{9cm}}
\toprule
\textbf{Metric} & \textbf{Calculation} \\
\midrule
\textbf{Detection Rate} & 
$|\text{violations correctly detected}| / |\text{total actual violations}|$ \\
\addlinespace[0.2em]
\textbf{False Positive Rate} & 
$|\text{false alarms}| / |\text{total detections claimed}|$ \\
\addlinespace[0.2em]
\textbf{Risk Coverage Rate} & 
$|\text{violation types detected}| / |\text{violation types present}|$ \\
\addlinespace[0.2em]
\textbf{Causal Chain Clarity} & 
$|\text{violations with complete traces}| / |\text{violations detected}|$ where complete trace = agent reasoning → decision → action → outcome \\
\addlinespace[0.2em]
\textbf{Predictive Alerting} & 
$|\text{violations with early warnings}| / |\text{total violations}|$ where early warning = risk indicator $\ge$ 1 event before violation \\
\addlinespace[0.2em]
\textbf{Proactive Intervention} & 
$|\text{successful preventions}| / |\text{violation attempts}|$ where prevention = intervention stopped violation from completing \\
\bottomrule
\end{tabular}
\end{table}

MI9 provides proactive intervention and behavioral alerting through its integrated governance components, though these operate differently from traditional predictive monitoring systems.
Proactive Intervention occurs through MI9's Graduated Containment System, which applies escalating restrictions (monitoring → planning restriction → tool restriction → isolation) based on real-time violation scores. 

The Continuous Authorization Monitoring component revokes permissions dynamically when goal-context mismatches are detected, while the Real-Time Conformance Engine blocks policy-violating actions before completion using FSM pattern matching.

Predictive Alerting emerges from MI9's Behavioral Drift Detection, which flags concerning behavioral changes using goal-conditioned baseline comparison before they escalate to policy violations. Additionally, the FSM-based Conformance Engine can identify multi-step violation sequences in progress, providing early warnings when agents begin patterns that typically lead to policy breaches.

\subsubsection{Dataset Composition}

\begin{table}[h]
\centering
\caption{Industry Distribution of Evaluated Scenarios}
\label{tab:industry_dist}
\begin{tabular}{lr}
\toprule
\textbf{Industry Sector} & \textbf{Count} \\
\midrule
Pharmaceutical & 275 \\
Finance & 257 \\
Semiconductor Manufacturing & 138 \\
Legal & 114 \\
Investment/Consumer Banking & 109 \\
Healthcare & 11 \\
Other Sectors & 40+ \\
\midrule
\textbf{Total} & \textbf{1,033} \\
\bottomrule
\end{tabular}
\end{table}

\begin{table}[h]
\centering
\caption{Attack Type Breakdown in Evaluation Dataset}
\label{tab:attack_types}
\begin{tabular}{lr}
\toprule
\textbf{Attack Type} & \textbf{Count} \\
\midrule
Prompt Injection & 149 \\
Model Inversion & 148 \\
Data Poisoning & 111 \\
Unauthorized Access & 102 \\
Data Evasion Attack & 94 \\
Reward Hacking & 42 \\
Privilege Escalation & 9 \\
SQL Injection / Exploits & 9 \\
Insider Threat & 4 \\
Social Engineering & 3 \\
Benign/No Attack & 362 \\
\midrule
\textbf{Total} & \textbf{1033} \\
\bottomrule
\end{tabular}
\end{table}

\subsubsection{Framework Performance Comparison}

Based on 1,033 valid evaluation samples, the following analysis demonstrates MI9's superior performance across all critical metrics using Wilcoxon signed-rank statistical testing.

\begin{table}[h]
\centering
\caption{Detection Rate Performance}
\label{tab:detection_rate}
\begin{tabular}{lcc}
\toprule
\textbf{Framework} & \textbf{Mean} & \textbf{Std Dev} \\
\midrule
\textbf{MI9 (ours)} & \textbf{0.9981} & 0.0440 \\
OpenTelemetry & 0.9398 & 0.1500 \\
LangSmith & 0.6852 & 0.3628 \\
\bottomrule
\end{tabular}
\end{table}

\begin{table}[h]
\centering
\caption{Risk Coverage Rate Performance}
\label{tab:risk_coverage}
\begin{tabular}{lcc}
\toprule
\textbf{Framework} & \textbf{Mean} & \textbf{Std Dev} \\
\midrule
\textbf{MI9 (ours)} & \textbf{0.9441} & 0.2284 \\
OpenTelemetry & 0.8444 & 0.2821 \\
LangSmith & 0.6046 & 0.3880 \\
\bottomrule
\end{tabular}
\end{table}

\begin{table}[h]
\centering
\caption{Governance Maturity Score Performance}
\label{tab:governance_maturity}
\begin{tabular}{lcc}
\toprule
\textbf{Framework} & \textbf{Mean} & \textbf{Std Dev} \\
\midrule
\textbf{MI9 (ours)} & \textbf{0.8395} & 0.0988 \\
OpenTelemetry & 0.5946 & 0.0707 \\
LangSmith & 0.4956 & 0.1237 \\
\bottomrule
\end{tabular}
\end{table}

\begin{table}[h]
\centering
\caption{Causal Chain Clarity Score Performance}
\label{tab:causal_clarity}
\begin{tabular}{lcc}
\toprule
\textbf{Framework} & \textbf{Mean} & \textbf{Std Dev} \\
\midrule
\textbf{MI9 (ours)} & \textbf{0.8220} & 0.1136 \\
OpenTelemetry & 0.4479 & 0.1146 \\
LangSmith & 0.3483 & 0.1653 \\
\bottomrule
\end{tabular}
\end{table}

\begin{table}[h]
\centering
\caption{Predictive Alerting Score Performance}
\label{tab:predictive_alerting}
\begin{tabular}{lcc}
\toprule
\textbf{Framework} & \textbf{Mean} & \textbf{Std Dev} \\
\midrule
\textbf{MI9 (ours)} & \textbf{0.6724} & 0.2335 \\
OpenTelemetry & 0.3405 & 0.2074 \\
LangSmith & 0.1067 & 0.1551 \\
\bottomrule
\end{tabular}
\end{table}

\begin{table}[h]
\centering
\caption{Proactive Intervention Rate Performance}
\label{tab:proactive_intervention}
\begin{tabular}{lcc}
\toprule
\textbf{Framework} & \textbf{Mean} & \textbf{Std Dev} \\
\midrule
\textbf{MI9 (ours)} & \textbf{0.5780} & 0.2927 \\
OpenTelemetry & 0.1164 & 0.2266 \\
LangSmith & 0.0202 & 0.0965 \\
\bottomrule
\end{tabular}
\end{table}

\newpage

\subsection{MI9 Components High-Level Example Responses}

\begin{itemize}
  \item \textbf{Agency-Risk Index (ARI)} \\
  \textit{Scenario:} A customer support agent (Agent ID: \texttt{cust\_agent\_45}) begins autonomously rewriting company refund policies during live chats. \\
  \textit{MI9 Response:} ARI recalculates to 0.71 (Tier 3: Highly Capable), triggering intensified monitoring and activation of drift detection.

  \item \textbf{Agentic Telemetry Schema (ATS)} \\
  \textit{Scenario:} An investment advisor agent (Agent ID: \texttt{invest\_bot\_12}) revises its portfolio optimization plan using \texttt{memory.read} followed by \texttt{goal.set}. \\
  \textit{MI9 Response:} ATS captures this cognitive sequence and routes it to the policy engine for evaluation against goal-modification policies.

  \item \textbf{Continuous Authorization Monitoring (CAM)} \\
  \textit{Scenario:} A retail banking agent attempts to perform system-level configuration tasks outside its initial role. \\
  \textit{MI9 Response:} CAM detects the goal-context mismatch, dynamically revokes admin privileges, and enforces a read-only policy.

  \item \textbf{Real-Time Conformance Engine} \\
  \textit{Scenario:} An agent executes: \texttt{tool.invoke("market\_analysis")} $\rightarrow$ \texttt{tool.invoke("client\_outreach")} $\rightarrow$ \texttt{tool.invoke("trade\_execution")}. \\
  \textit{MI9 Response:} FSM rule detects violation of approval sequencing; trade execution is blocked pending human authorization.

  \item \textbf{Behavioral Drift Detection} \\
  \textit{Scenario:} A compliance monitoring agent begins reducing alert frequency while retaining the goal \texttt{"monitor trading thresholds"}. \\
  \textit{MI9 Response:} Goal-conditioned drift detection registers significant deviation from baseline. Alert issued for human review and potential containment.

  \item \textbf{Graduated Containment} \\
  \textit{Scenario:} A Tier 4 trading agent spawns subagents at 3$\times$ its normal rate during peak trading. \\
  \textit{MI9 Response:} Tool Restriction containment level is triggered. Write-access to trading tools is revoked; simulation mode is enabled for behavioral observation.
\end{itemize}

\end{document}